# Panchromatic Sharpening of Remote Sensing Images Using a Multi-scale Approach

Hamid Reza Shahdoosti

***Abstract*— An ideal fusion method preserves the Spectral information in fused image and adds spatial information to it with no spectral distortion. Recently wavelet kalman filter method is proposed which uses ARSIS concept to fuses MS and PAN images. This method is applied in a multiscale version, i.e. the variable index is scale instead of time. With the aim of fusion we present a more detailed study on this model and discuss about rationality of its assumptions such as first order markov model and Gaussian distribution of the posterior density. Finally, we propose a method using wavelet Kalman Particle filter to improve the spectral and spatial quality of the fused image. We show that our model is more consistent with natural MS and PAN images. Visual and statistical analyzes show that the proposed algorithm clearly improves the fusion quality in terms of: correlation coefficient, ERGAS, UIQI, and Q4; compared to other methods including IHS, HMP, PCA, A`trous, udWI, udWPC, Adaptive IHS, Improved Adaptive PCA and wavelet kalman filter.**

***Index Terms*— A`trous wavelet transform, image-fusion, Kalman particle filter, multi-scale analysis.**

## I. Introduction

In remote sensing systems, scenes are observed in different portions of electromagnetic spectrum; therefore, the remote sensing images vary in spectral and spatial resolution. To collect more photons and maintain signal to noise ratio simultaneously, the multispectral sensors, with high spectral resolution, have a lower spatial resolution compared to panchromatic image which has a higher spatial resolution, and wide spectral bandwidth. With proper algorithms it is possible to fuse these images and produce imagery with the best information of both, namely high spatial and high spectral resolution. An appropriate image fusion method should minimize ambiguity and redundancy in the output while maximizing the relative information specific to an application [1].

Existing algorithms for the fusion are classified generally into three categories: 1- projection and component substitution algorithms, such as IHS, GS (Gram Schmidt), PCA (Principal Component Analysis) and CA (Correspondence Analysis) fusion [2]-[5]. 2- fusion methods based on Band ratio and arithmetic combination, such as Brovey, SVR (Synthetic Variable Ratio) , HPM (High Pass Modulation) and RE (Ratio Enhancement) [6]. 3- Multi-resolution fusion algorithms, which inject spatial information from the high-resolution images into the multispectral images, such as wavelets, Gaussian Laplacian pyramid techniques [7]–[9].

The early image-fusion methods are based on the IHS (also known as HSI) [3], principal component analysis (PCA) [10], [11], HPF (High Pass Filtering) [12] and HPM (High Pass modulation) [13], [14]. But these methods can cause spectral distortion in the results. This distortion of the spectral information during the fusion process is not acceptable in most applications, such as land uses/cover mapping though a spectral classification procedure.

The emergence of methods based on multi-resolution analysis significantly improved the results of the image fusion. These algorithms are used to perform a hierarchical description of the information content relative to spatial structures in an image. The well-known Mallat's fusion algorithm [15] uses an orthonormal basis and can effectively preserve spectral information, but because of down-sampling operators, which are used to implement ordinary discrete wavelet transforms, these transforms are not shift-invariant and can lead to problem in data fusion [16].

To avoid above-mentioned problem, the discrete wavelet transform known as "à trous" algorithm [9] was proposed by eliminating the decimation operators in wavelet structure. It is a shift-invariant and redundant wavelet transform algorithm based on a multiresolution dyadic scheme. The ATWT (A`trous Wavelet Transform) allows an image to be decomposed into nearly disjointed bandpass channels in the spatial frequency domain and preserves the spatial connectivity of its highpass details, e.g., textures and edges.

As motioned above, the major drawback of Projection–Substitution algorithms such as PCA and IHS, is spectral distortion, which is distinguished by a trend to present a predominance of a color on the others [17]. These methods assume that Pan image is equal to a linear combination of LMS images. As a matter of fact, this presumption is not realistic because LMS airborne or spaceborne sensors do not have a constant response over the whole bandwidth. In addition, local dissimilarities are not taken into account in these methods. Because of the local dissimilarities such as contrast inversions and occultations of objects, the correlation between the images decreases, and the images are not correctly synthesized by these fusion methods [17]. Some new proposed Projection–Substitution algorithms such as improved adaptive

PCA [18] and adaptive HIS [19] eliminate this problem to some extent by partial injection of PAN image into the LMS images.

Band ratio and arithmetic combination methods such as Brovey and HPM can be obtained by modeling of optical remote sensing sensors. However, the result of these approaches is not suitable because image formation in reality is far more complex than the simplified image formation model [14].

Multi-resolution fusion algorithms have demonstrated a superior capability of translating the spatial information of the finer scale PAN data to the coarser scale of the LMS images with minimal introduction of spectral distortions. The missed high frequency components of LMS images are located in the high frequency components of the PAN image. To improve the spatial resolution of LMS images only this part is needed. This method preserves the lower frequency (spectral information) of the LMS images which makes reduction in the spectral distortion of the fused images. However, all of the high frequency components of the PAN image do not belong to each LMS images, Hence, in order to improve the quality of the fused image, we should apply a transformation to convert the information provided by the multiscale representation of the PAN image into the information needed for the synthesis of the MS images (multispectral images) [20].

The ARSIS concept, from its French name "Amélioration de la Résolution Spatiale par Injection de Structures" (improvement of spatial resolution by structure injection), makes use of a Multi-resolution method for modeling and description of the missing information between the PAN images and MS image. Several investigations have showed that the best presently achievable results are obtained by the methods based on the ARSIS concept [21].

Recently wavelet Kalman filter method is proposed which uses ARSIS concept to fuses images [22]. Because the injection is conducted on spatial details extracted by an ATWT decomposition, the Kalman filter is applied in a multiscale version, i.e. our variable index is scale. By Appling the wavelet Kalman filter, which uses a predictive model, we can inject spatial details of the PAN image accurately. We present a more detailed study on this model and discuss about rationality of its assumptions. Finally, we propose a new algorithm to improve the spectral and spatial quality of the fused image. We call this method wavelet Kalman particle filter. The experimental results show that the proposed algorithm significantly improves the fusion quality in terms of RASE, ERGAS, SAM, correlation coefficient, and UIQI compared to other fusion methods. The paper is structured in six sections. In Section II we present the coefficient relationships of undecimated wavelet and model the image fusion problem. Section III is devoted to the Kalman particle filter. Section IV proposes image fusion based on wavelet kalman particle filtering. The experimental results and discussions are described in Section V. Finally, our conclusions are given in Section VI.

## II. Coefficient Relationships

Reference [23] and [24] showed that contourlet and decimated wavelet coefficients of natural images are *approximately* decorrelated but dependent on each other since conditional expectations is approximately zero and good mutual information exist for general natural images.

We want to study the statistics of the udW (Undecimated Wavelet) coefficients of natural images in the first step. As depicted in Fig. 1, some important undecimated wavelet coefficient relationships are defined. For each udW coefficient $X$, we define the coefficients in the same spatial location in the coarser scales as its predecessors ($PX_i$), where $PX_1$ and $PX_2$ are its parents and grandparents respectively.

To study the statistical model, we experiment various natural images of size 512 × 512. Both images with simple edge-dominant images such as “Peppers” and textured images such as “Barbara” are used in our experiments. The 9-7 biorthogonal filters (referred to as 9-7 filters) is used for the multiscale decomposition.

Fig. 2 shows the conditional distributions of udW coefficients, conditioned on their parents ($PX_1$) and grandparents ($PX_2$), using the “Barbara” image. In fact this shows that there is correlation between $X$ and $PX_1$, even though $X$ and $PX_2$ are approximately decorrelated since conditional expectations $E[X/PX_2]\approx 0$.The $X$ and $PX_1$ are uncorrelated in decimated Wavelet decomposition. However, because of the redundancy of the information, correlation between $X$ and $PX_1$ in udW can be seen

To investigate the validity of the first order markov model of the udW, we should estimate mutual information between $X$ and its predecessors. Mutual information can be interpreted as how much information random variables contain about each other. Mutual information increases with increasing dependence between the two variables. To estimate the mutual information between udW coefficient and its parents, we use the following estimator [25]:

$$\hat{I}(X;Y)=\sum_{i,j}\frac{k_{ij}}{N}\log\frac{k_{ij}N}{k_i k_j}-\frac{(J-1)(K-1)}{2N} \tag{1}$$

where $k_{ij}$ is the number of sample in the joint histogram cell $(i,j)$. $k_i=\sum_j k_{ij}$ and $k_j=\sum_i k_{ij}$ are the marginal distribution histogram, N is the total number of samples, and J and K are the number of histogram cells along $X$ and $Y$ variable respectively. J and K are so chosen to give the maximum estimate in (1).

The first term in (1) is the mutual information histogram estimate and the second term is a partial bias correction term.

Note that $X$, $PX_1$, $PX_2$,…, $PX_n$ is a Markov chain if and only if $X$ and $PX_n$, are conditionally independent given $PX_{n-l}$ (l>1), or in other words, I($X$:$PX_1$, $PX_2$,…, $PX_i$)= I($X$:$PX_1$) (i=1,…n).

For mutual information with a large set of variables, the estimator in (1) becomes inaccurate. For example, consider estimating the mutual information I($X$:$PX_1$, $PX_2$,…, $PX_i$) between udW coefficients, as the number of $i$ increases, mutual information estimation accuracy decreases exponentially [26]. In such cases, we use the sufficient statistic instead of the high dimensional variable $\{PX_i\}_{i=1}^{n}$ as follows [26]:

$$T = \sum_{i=1}^{n} a_i \left|PX_i\right| \tag{2}$$

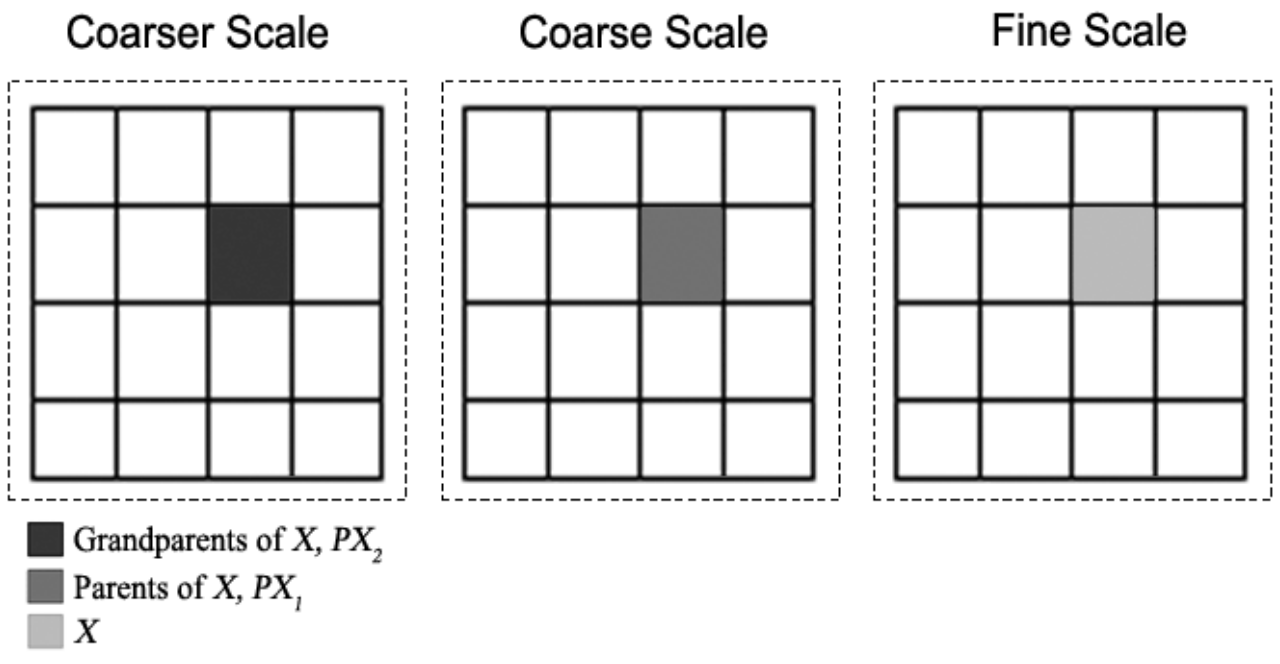

Fig. 1. udW coefficient relationships.

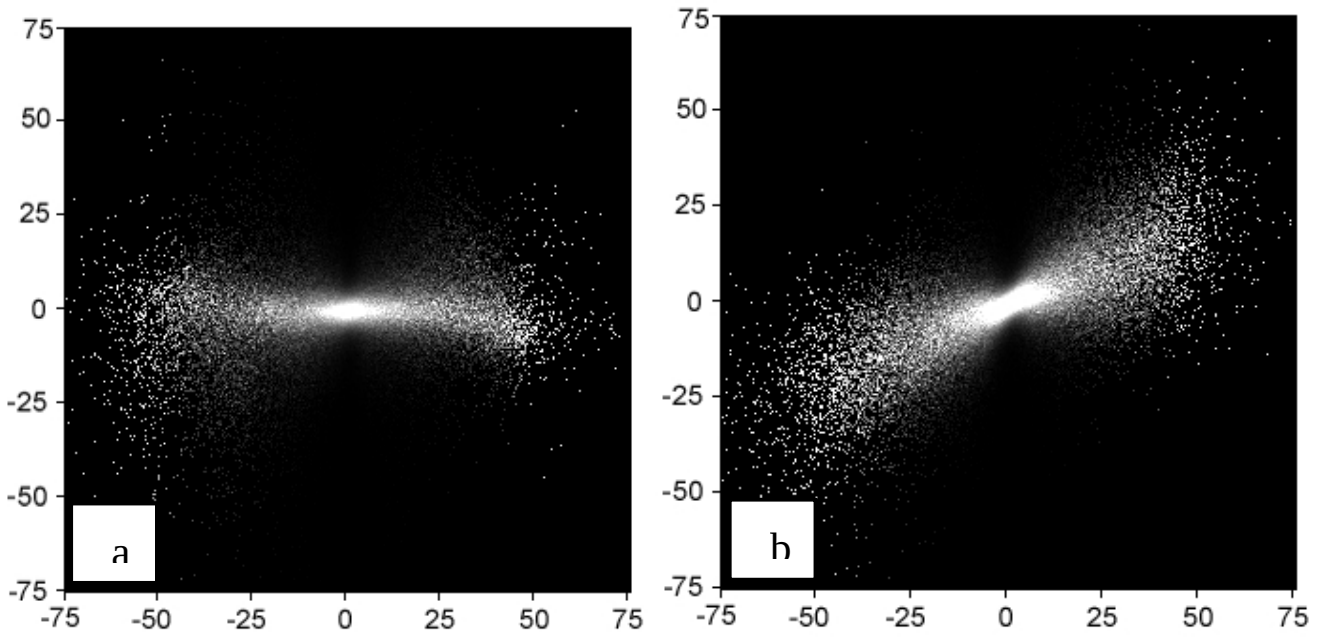

Fig. 2. conditional distributions of udW coefficients, conditioned on (a) parents (*PX*$_1$) (b) grandparents (*PX*$_2$) for the "Barbara" image.

Where $a_i$ are so chosen as to maximize I(*X;T*), using standard optimization algorithms in MATLAB. Then I(*X;T*) have only two variables and can be estimated accurately by (1). I($X{:}PX_1, PX_2, \ldots, PX_i$), (i=1,…n) is shown in Fig. 3. As seen in Fig. 3, udW coefficients obey the first order markov model approximately, so we can model it by the state-space model described in the next section.

## III. Kalman Particle Filtering

Unlike the standard Kalman Filter that makes a Gaussian assumption to simplify the optimal Bayesian estimation the particle filter which is a sequential Monte Carlo method does not [27]. So when the mentioned assumption doesn't match, particle filter follows the more optimum way to estimate than kalman filter. According to state-space model we assume that the true state at time $k$ is related to the state at $(k-1)$ as follows:

$$s_k = A_k s_{k-1} + u_k \tag{3}$$

Where, as showed in previous section this assumption is approximately rational. In addition, this assumes that at time $k$ an observation $x_k$ of the true state $s_k$ is made as below:

$$x_k = H_k s_k + w_k \tag{4}$$

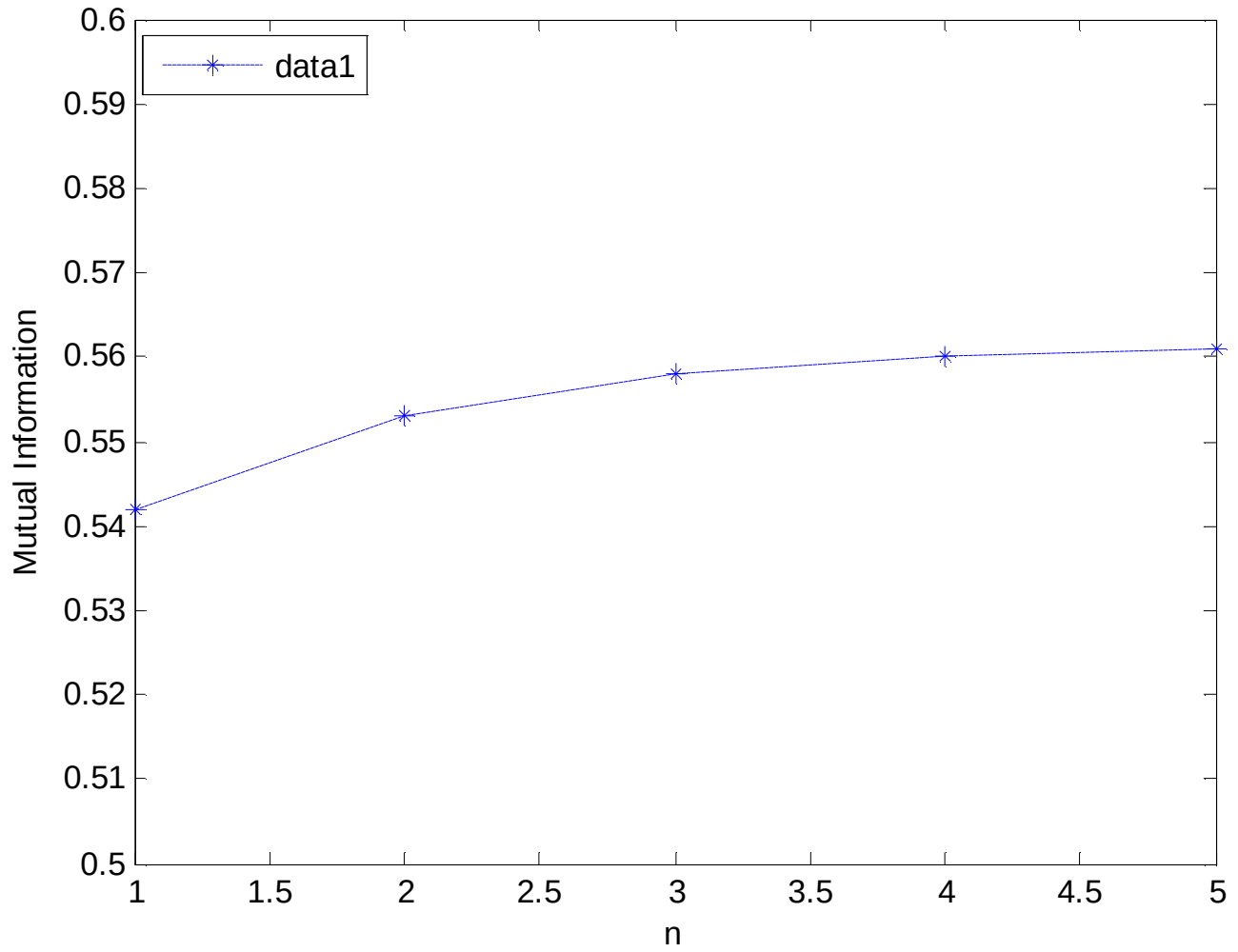

Fig. 3. I($X{:}PX_1, PX_2, \ldots, PX_i$) (i=1,…5) curve for "Barbara" by using its sufficient statistic.

Equation 3 implies that state vector can be considered as a Gauss-Markov model in that $u_k$ is the process noise which is a zero mean AWGN noise with covariance matrix $R^u$. And equation 4 expresses that observation vector can be depicted as a linear combination of state and an error $w_k$ which is a zero mean AWGN noise with covariance matrix $R^w$. Besides the initial state, and the noise vectors at each step, $\{s_0, w_1, \ldots, w_k, v_1, \ldots, v_k\}$ are all assumed to be mutually independent.

Aiming to estimate the state $s_k$, particle filter recursively achieves an approximated representation of the posterior state distribution[1] using sequential importance sampling (SIS) [28]. Therefore as mentioned, particle filter is a method to estimate the posterior density based on Monte Carlo simulation and SIS. In Monte Carlo simulation, the goal is to approximate the posterior density by a set of samples (called as particle), drawn from its distribution as follows

$$\hat{P}(s_k/\chi_k) = \frac{1}{N}\sum_{i=1}^{N} d(s_k - s_k^{(i)}) \tag{5}$$

Where $\chi_k$ is denoted as a set of observation vectors, $\chi_k = \{x_0, \ldots, x_k\}$, the particles $\{s_k^{(i)}\}_{i=1}^{N}$ are drawn from the posterior $P(s_k/\chi_k)$ and $N$ is the number of drawn particles. Therefore, expectations of any desired function could be approximated as below

$$E(f(s_k)) = \int f(s_k) P(s_k/\chi_k) ds_k \approx \frac{1}{N}\sum_{i=1}^{N} f(s_k^{(i)}) \tag{6}$$

But it is often impossible to draw (or sample) from the posterior density, as mentioned in [29] it is common to evade this problem by sampling from a known distribution, so-called proposal distribution denoted as $q(s_n/\chi_n)$, hence equation (6) can be rewritten as [29]:

[1]Let $\chi_k$ be as a set of observation vector, $\chi_k = \{x_0, \ldots, x_k\}$, then the posterior state distribution(or posterior density) is defined as $P(s_k/\chi_k)$.

$$E(f(s_k)) = \frac{E_{q(s_k/\chi_k)}(W_k(s_k)f(s_k))}{E_{q(s_k/\chi_k)}(W_k(s_k))} \tag{7}$$

Where $\{W_k(s_k)\}_{i=1}^{N}$ are so-called importance weights and are defined as:

$$W_k(s_k) = \frac{P(\chi_k/s_k)P(s_k)}{q(s_k/\chi_k)} \tag{8}$$

Consequently by drawing the particles $\{s_k^{(i)}\}_{i=1}^{N}$ from the proposal distribution $q(s_k/\chi_k)$ equation (7) is approximated by

$$E(f(s_k)) = \sum_{i=1}^{N} \tilde{W}_k(s_k^{(i)})f(s_k^{(i)}) \tag{9}$$

Where

$$\tilde{W}_k(s_k^{(i)}) = \frac{W_k(s_k^{(i)})}{\sum_{j=1}^{N} W_k(s_k^{(j)})} \tag{10}$$

And if the proposal distribution can be factorized as below

$$q(S_k/\chi_k) = q(S_{k-1}/\chi_{k-1})q(s_k/S_{k-1},\chi_k) \tag{11}$$

The importance weights can be updated recursively:

$$W(s_k^{(i)}) = W(s_{k-1}^{(i)})\frac{P(x_k/s_k^{(i)})P(s_k^{(i)}/s_{k-1}^{(i)})}{q(s_k^{(i)}/S_{k-1}^{(i)},\chi_k)} \tag{12}$$

Where the $S$ as $\chi$ is the set of state vectors, $S_k = \{s_0,...,s_k\}$.The serious problem of the sequential importance sampling mentioned so far is increasing of the importance weights variance over time [30]. This phenomenon which is called the degeneracy can be solved by resampling methods [29] that suppress particles with low weights and multiply particles with high weights. Many resampling algorithm has been developed to circumvent the degeneracy but we used the residual resampling which make less variance than well-known ones [31]. The particle filter which is used in our fusion algorithm is summarized in Table I. As can be seen the proposal distribution should be determined, in fact the choice of proposal distribution is a critical issue in particle filtering; the optimal proposal distribution in the sense of minimizing the importance weights variance is as follows [32]:

$$q(s_k/S_{k-1},\chi_k) = P(s_k/S_{k-1},\chi_k) \tag{13}$$

But sampling from this distribution is almost impractical; a popular method is based on Gaussian approximation to this optimal proposal distribution by using a Kalman filter [29]. That is, at iteration *k* we use the Kalman filter to compute the mean and covariance of this Gaussian approximated proposal distribution for each particle with the aid of particle at previous iteration *k-1*:

$$q(s_n/S_{n-1},\chi_n) = \mathrm{N}(\bar{s}_k^{(i)},\hat{P}_k^{(i)}) \tag{14}$$

TABLE I
PARTICLE FILTER

Initialization: k=0
For i=1,…,N
draw the particles $s_0^{(i)}$ from the prior density $P(s_0)$
end
For k=1,2,…
Importance sampling step:
For i=1,…,N
Sample $\hat{s}_k^{(i)} \sim q(s_k^{(i)}/S_{k-1}^{(i)},\chi_k)$
end
For i=1,…,N
evaluate the importance weights:

$$w_k^{(i)} = w_{k-1}^{(i)}\frac{P(x_k/\hat{s}_k^{(i)})P(\hat{s}_k^{(i)}/s_{k-1}^{(i)})}{q(\hat{s}_k^{(i)}/S_{k-1}^{(i)},\chi_k)}$$

end
For i=1,…,N
Normalize the importance weights:

$$\tilde{w}_k^{(i)} = \frac{w_k^{(i)}}{\sum_{j=1}^{N} w_{k-1}^{(j)}}$$

end
Resampling step:
Eliminate/Multiply particles $(\hat{s}_k^{(i)},\hat{P}_k^{(i)})$ according to its normalized weights $\tilde{w}_k^{(i)}$ to obtain particles $(s_k^{(i)},P_k^{(i)})$.
Output, the optimal MMSE estimator is given as:

$$\hat{s}_k = E(s_k/\chi_k) \approx \frac{1}{N}\sum_{i=1}^{N} s_k^{(i)}$$

end

Which $\bar{s}_k^{(i)}$ and $\hat{P}_k^{(i)}$ are mean and covariance propagated by Kalman Filter for each particle. Finally this importance sampling procedure with mentioned proposal distribution is summarized in Table II.

TABLE II
IMPORTANCE SAMPLING WITH KALMAN FILTER

The importance sampling step in the particle filter,
For i=1,…,N, Sample $\hat{s}_k^{(i)} \sim q(s_k^{(i)}/S_{k-1}^{(i)},\chi_k)$
is replaced with the following kalman filter update:

For i=1,…,N
Update the particles by the kalman filter:

$$\bar{s}_{k/k-1}^{(i)} = A_k s_{k-1}^{(i)}$$
$$P_{k/k-1}^{(i)} = H_k P_{k-1}^{(i)} H_k^T + R_k^u$$
$$K_k^{(i)} = P_{k/k-1}^{(i)} H_k^T (R_k^w + H_k P_{k/k-1}^{(i)} H_k^T)^{-1}$$
$$\bar{s}_k^{(i)} = \bar{s}_{k/k-1}^{(i)} + K_k^{(i)}(x_k - H_k \bar{s}_{k/k-1}^{(i)})$$
$$\hat{P}_k^{(i)} = P_{k/k-1}^{(i)} - K_k^{(i)} H_k P_{k/k-1}^{(i)}$$

Sample $\hat{s}_k^{(i)} \sim q(s_k^{(i)}/S_{k-1}^{(i)},\chi_k) \approx \mathrm{N}(\bar{s}_k^{(i)},\hat{P}_k^{(i)})$
End

## IV. IMAGE FUSION BASED ON WAVELET KALMAN PARTICLE FILTERING

The aim is to estimate the missed high frequency components of MS images from the PAN by modeling the problem in the coarser resolution. To do this, we decompose LMS images and the PAN image into coarse to fine resolution levels by ATWT. Our variable index $k$ in kalman particle filtering is scale instead of time, so one step increasing of independent variable $k$ modeled as transition from coarser level to finer one. Fig. 4 shows the flow diagram of the wavelet kalman particle filter algorithm suitable for fusion of LMS and PAN images with a resolution ratio equal to 4, where, the state vector is the wavelet coefficients of LMS images, and observation vector is the wavelet coefficients of the PAN image. The kalman filter method for image fusion proposed in [22] gives a good estimate of desired states but because the posterior density is not necessarily Gaussian, We used the kalman particle filter that gives an approximated complete representation of posterior density to obtain a better estimation. In fact we approximated posterior density for k=2 and k=3 by the method reduced sufficient statistics proposed in [33] for some LMS and PAN images and found that the kurtosis of these distribution is considerably more than 3, hence they're not Gaussian. The coarser information of LMS can be used to correct and initialize the kalman particle filter. The different local radiometry and geometry between the PAN and LMS images clearly explained in [34]. In other words, the local high frequency spatial information that is visible in the LMS image can be missing in the PAN image, and vice versa. Therefore, if we substitute the estimated finer resolution components of the PAN image, spectral and spatial information of the LMS image may be eliminated. In addition, if we add the estimated finer resolution components of the PAN image to the LMS, redundant geometric information can be produced. This implies that in the fusion process, the LMS image must be maintained, and the high frequency, which is not present in the LMS image, must be injected. So, it is necessary to develop an effective fusion algorithm that injects the necessary spatial information and preserves the high frequency of LMS image. So the finer resolution components of the LMS images can be considered as another observation vector. We called this finer resolution HDMS (High Frequency Components of Degraded Multi Spectral image). Exploiting of this information with the aim of image fusion is not for first time, in fact the superiority of Additive Wavelet method [9] compared to Substitution Wavelet method is because of its HDMS preserving in the fusion result.

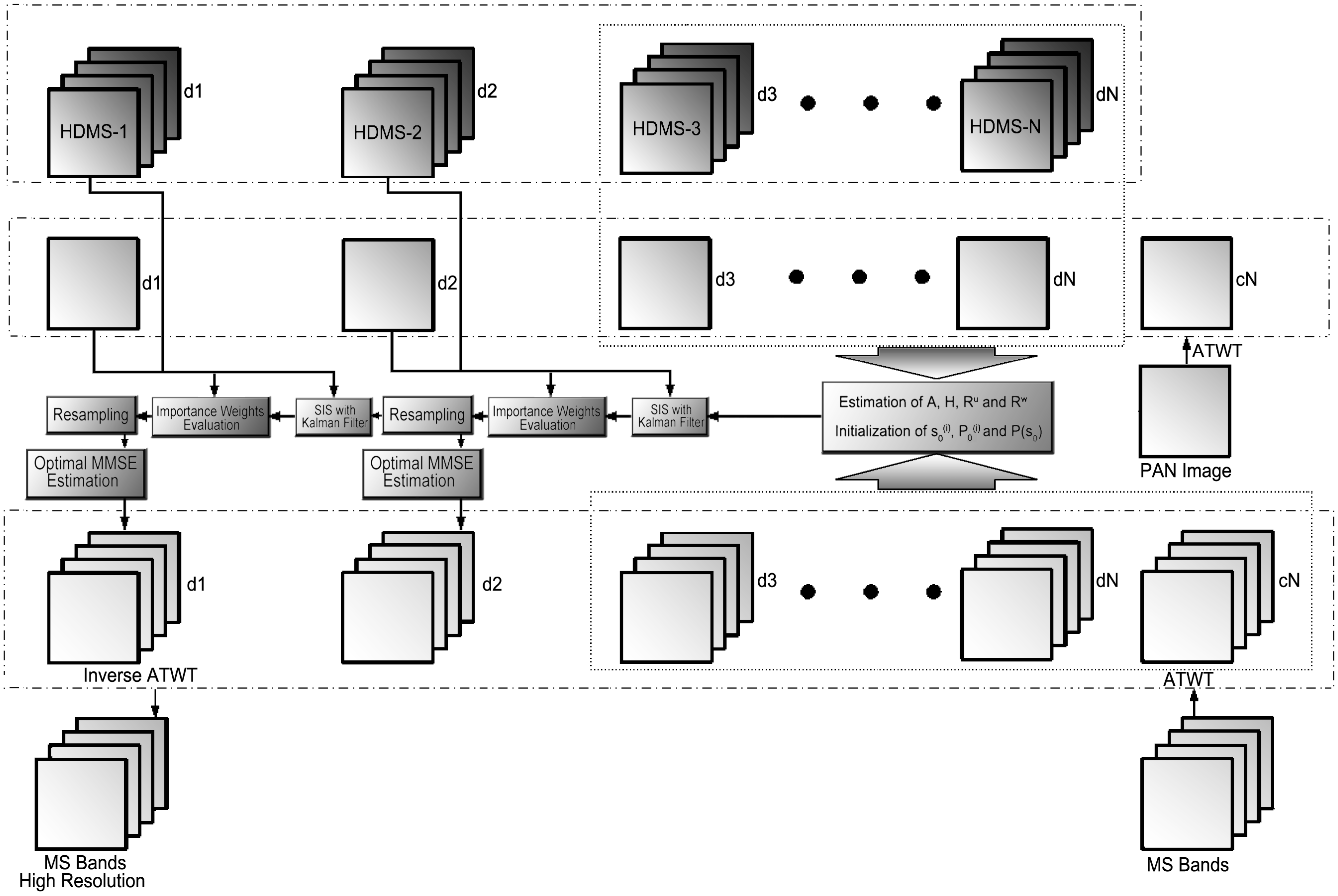

Fig. 4. Wavelet-Kalman Particle filter algorithm applied for resolution ratio = 4.

Therefore in addition to coarser resolution of PAN image (which is shown in Fig 4), we need coarser HDMS to initialize the kalman particle filter. We can produce these HDMSes by degrading the LMS. For each step, we need an observation vector of HDMS, so we degrade LMS spatially corresponding to the required level to obtain HDMS for each level (see Fig. 4). By this approach, we benefit of both fine resolution components of the PAN image and high frequency components of LMS image in estimation of High frequency components of MS images. Therefore, finer resolution details of MS images are estimated from the coarser resolution details of MS images and the set of finer resolution details of PAN and HDMS with the aid of importance weights. To show the dependency between HDMS and higher frequency components of MS images, we degrade "lena" one level and plot the highest frequency of "lena" versus the highest frequency of degraded "lena". Fig. 5 shows correlation between these two components and verifies the idea of using HDMS for estimation of higher frequency components in the natural images.

The $A$ (transition matrix) and $R^u$ (covariance matrix of process noise) can be approximated by Yule-Walker equation according to AR(1) model in coarser levels in which the true wavelet coefficients of MS are available.

To estimate H (measurement matrix), we follow the method in [19]:

$$dP_i \approx \sum_{j=1}^{K} a_{ij} dM_{ij}, \quad i = 2,...,N \tag{15}$$

$$dHDMS_{ij} \approx a_{ij} dM_{ij}, \; i = 2,...,N \,;\, j = 2,...,K \tag{16}$$

In order to calculate these coefficients we created the following function G to minimize with respect to $a_j$ :

$$\min G(a_i) = \sum_x (\sum_j (a_{ij} dM_{ij}) - dP_i)^2 + \gamma \sum_j (\max(0, -a_{ij})))^2 \tag{17}$$

$$\min G(a_j) = \sum_x (a_{ij} dM_{ij} - dHDMS_{ij})^2 + \gamma(\max(0,\text{-}a_{ij}))^2 \tag{18}$$

$$, i = 2,...,N \,;\, j = 2,...,K$$

The first term ensures that the coefficients yield a linear combination that approximates the panchromatic image. Physically we do not want negative $a_j$ , therefore by using the Lagrange multiplier $\gamma$ we add the non-negativity constraint on the $a_j$ . This equation can be solved by discretizing the PDE to obtain a semi-implicit scheme as represented in [19]. We compute $a_j$ factors in each level (i) for the set of observations (details coefficient of the PAN image and HDMS) separately. $a_j$ factors of the final levels for two observations, where the true wavelet coefficients of MS are not available can be estimated from previous $a_j$ factors with the aid of curve fitting. We found experimentally that the measurement noise distribution in this method is more similar to Gaussian distribution (kurtosis of observation noise is closer to 3) than method proposed in [22]. Therefore with normality assumption of measurement noise kalman particle filter follows the optimal way to estimate the state vector.

$R^w$ (observation noise) is the mean square error between available MS coefficients and two observation vectors. As it is obvious, this matrix is not diagonal because of redundancy information of finer resolution details of PAN and HDMS.

Finally It's worth to note that $P(s_0)$, $s_0^{(i)}$ and $P_0^{(i)}$ should be determined as the initialization set of kalman particle filter, As illustrated in Fig. 4 these can be obtained by applying kalman particle filter in coarser resolution in which the true states are available.

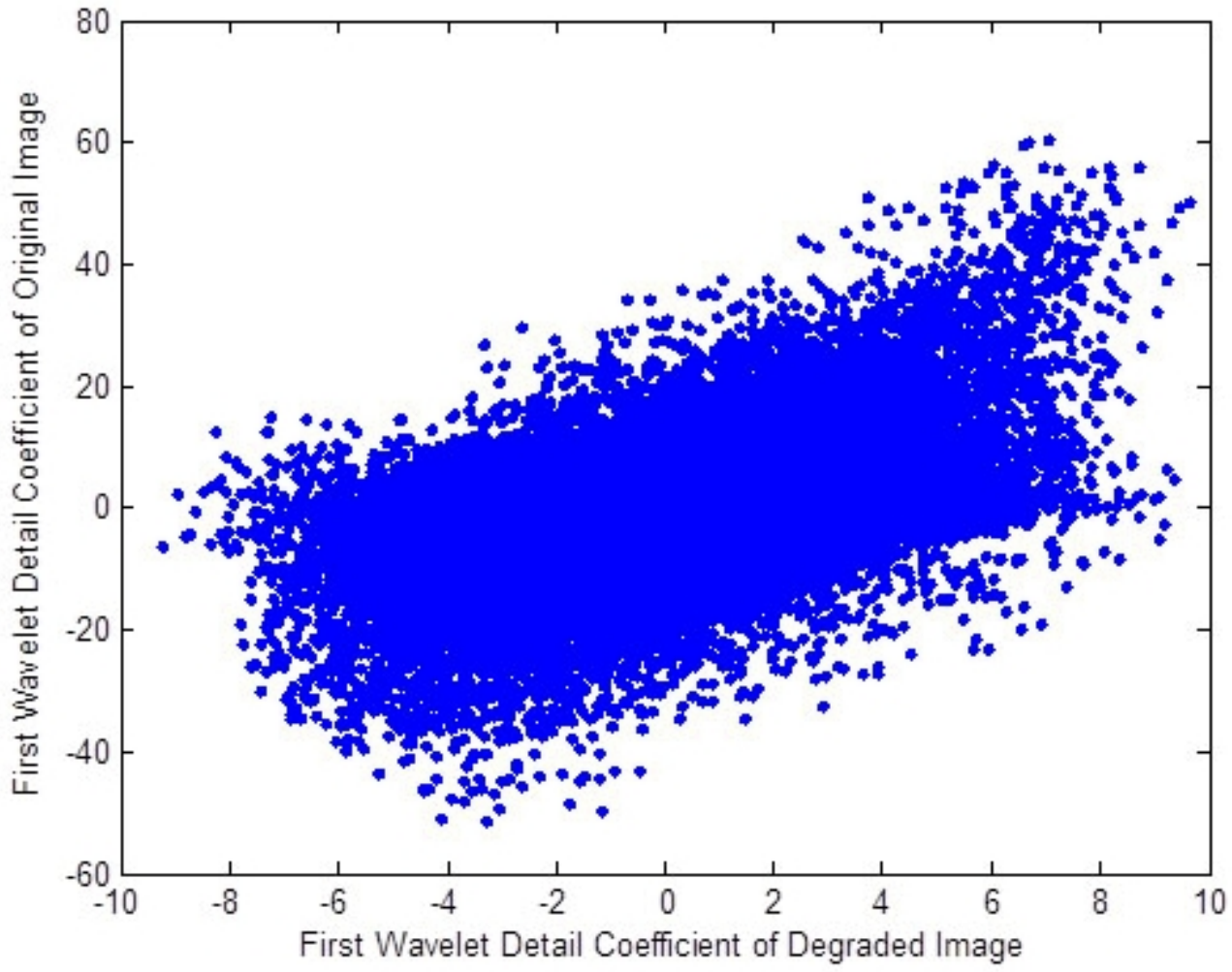

Fig. 5. Highest frequency component versus HDMS for "Lena".

## V. Experimental Results and Discussion

Four-band multispectral QuickBird data was used for our experiments. These images are acquired by a commercial satellite, QuickBird. The QuickBird data set was taken over the Pyramid area of Egypt in 2002, which collects one 0.7 m resolution panchromatic band (450–900 nm) and blue (450–520 nm), green (520–600 nm), red (630–690 nm), near infrared (760–900 nm) bands of 2.8 m resolution. The subscenes of the raw images are used as PAN and LMS.

### A. *Application to Full Resolution*

Visual comparison of the fused images is the first step of quality assessment. To do this, several aspects of the image quality such as surfaces, linear features, edges, color, blurring, and blooming can be taken into account. The visual performances of the Quickbird data are shown in Fig. 6 for an RGB composition.

We selected wavelet kalman filter, Improved adaptive PCA and adaptive IHS methods for visual assessment. Although all of these new methods lead to the good result, we just mention some deficiencies of these algorithms relative to our method here .As represented in Fig. 6(d), some of small objects such as blue busses miss their spectral content in Wavelet Kalman Filter method, because this algorithm doesn't consider HDMS in its fusion process. As illustrated in Fig. 6(e), some false edges can be seen in Adaptive IHS at the bottom of the Images. These artifacts probably come from unsuitable parameters of the edge detector (see [19]). The colors of some objects such as blue busses are lost in Adaptive IHS too. The geometry enhancement has been pretty well in Improved Adaptive PCA method, but some colors have been predominated on the others. As can be seen from Fig. 6(c), Wavelet Kalman particle filter method provides merged images of higher spectral quality than the Wavelet Kalman

Filter, Adaptive IHS and Improved Adaptive PCA.

In addition to the visual inspection, the performance of each method should be analyzed quantitatively. Recently proposed by Alparone *et al.* [35], the QNR index evaluates the quality of the fused image without requiring the reference MS image and combines the two distortion indexes of the radiometric and geometric distortion indexes.

Table III shows the performance comparisons of the fused images by QNR index. This Table shows that an image with simultaneous high spectral and spatial resolution can be achieved by the proposed method.

### B. Application to an Inferior Level

Although studies show that the quality of the synthesized images at the high resolution level cannot be estimated from the assessments at the inferior level, nonetheless it seems rational to assume that the quality of the synthesized images at the high resolution level is close to the quality of the synthesized images at the inferior level [14].

Visual assessment is easier at this level, because the reference image is available. Fig. 7 Shows degraded image, reference image, and fusion result of Wavelet Kalman Particle filter, Wavelet Kalman Filter, Adaptive IHS and Improved Adaptive PCA. This figure illustrates larger area of the raw image because it is more proper for visual assessment.

As can be seen from Fig. 7(c), (d), in most area of the image, the colors are closer to the reference image in Wavelet Kalman filter and Wavelet Kalman Particle filter methods. However, geometric information of Wavelet Kalman filter is not sufficient and the result is relatively blurred. False edges of Adaptive IHS at Pyramid area are obvious [see Fig. 7(e)]. Sharpening of Improved Adaptive PCA is more than desirable and colors have been changed during the fusion process which presents the spectral distortion, [see Fig. 7(f)]. As illustrated in Fig. 7 (d), the result of proposed method preserves both spectral and spatial information of the PAN and MS image simultaneously.

In addition to the visual inspection, the performance of each method should be analyzed quantitatively. In order to assess the quality of the merged images at the inferior level, five objective indicators were used.

#### 1) Correlation coefficient

This metric indicates the degree of linear dependence between the original reference and fused images [9]. If two images are identical, the correlation coefficient will be maximal and equals 1. It is defined as follows:

$$Corr(A,B)=\frac{\sum_{j=1}^{npix}(A_j-m_A)(B_j-m_B)}{\sqrt{\sum_{j=1}^{npix}(A_j-m_A)\sum_{j=1}^{npix}(B_j-m_B)}} \tag{21}$$

#### 2) ERGAS

ERGAS (or relative global dimensional synthesis error) is as follows [2]:

$$ERGAS=100\frac{h}{l}\sqrt{\frac{1}{N}\sum_{i=1}^{N}\frac{RMSE^2(B_i)}{M_i^2}} \tag{22}$$

Where $h$ is the resolution of the high spatial resolution image, $l$ *is* the resolution of the low spatial resolution image, and $M_i$ is the mean radiance of a specific band involved in the fusion.

TABLE III: QNR INDEX FOR FIG. 6

| | $D_l$ | $D_S$ | QNR |
|---|---|---|---|
| **Adaptive IHS** | 0.0238 | 0.0534 | 0.9466 |
| **Improved Adaptive PCA** | 0 .0491 | 0.0310 | 0.9690 |
| **Wavelet Kalman filter** | 0.0201 | 0.0517 | 0.9483 |
| **Proposed Method** | **0.0183** | **0.0288** | **0.9712** |

RMSE is the root mean square error and can be computed by the following expression:

$$RMSE^2(B_i)=Bias^2(B_i)+SD^2(B_i) \tag{23}$$

The lower the value of the ERGAS is, the higher the quality of the merged images is.

#### 3) UIQI

It is defined as follows [35]:

$$Q=\frac{s_{AB}}{s_A s_B}\frac{2m_A m_B}{m_A^2+m_B^2}\frac{2s_A s_B}{s_A^2+s_B^2} \tag{24}$$

Where $A$ and $B$ are fused and reference images respectively. The universal image quality index (UIQI) models any distortion as a combination of three different factors: loss of linear correlation, contrast distortion and luminance distortion.

#### 4) $Q_4$

$Q_4$ is composed of three different factors [37]: The first is the modulus of the hyper-complex CC between the two spectral pixel vectors and it is sensitive to both the loss of correlation and the spectral distortion between the two MS data sets. Contrast changes and mean bias on all bands are measured by the second and third factors, respectively

#### 5) SA or SAM

The spectral angle mapper (SA or SAM) between each band of the original reference multispectral and the fused image is calculated as follow [35]:

$$SAM(\mathrm{u},\hat{\mathrm{u}})=\arccos\left(\frac{\langle \mathrm{u},\hat{\mathrm{u}}\rangle}{\|\mathrm{u}\|_2\|\hat{\mathrm{u}}\|_2}\right) \tag{25}$$

Where $u=\{u_1,u_2,...,u_l\}$, the original reference spectral pixel is vector, and $\hat{u}=\{\hat{u}_1,\hat{u}_2,...,\hat{u}_l\}$ is the spectral pixel vector of the fused image. SAM is calculated in degrees for each pixel vector, and it is averaged on all pixels to obtain a single value. It should be as close to 0 as possible.

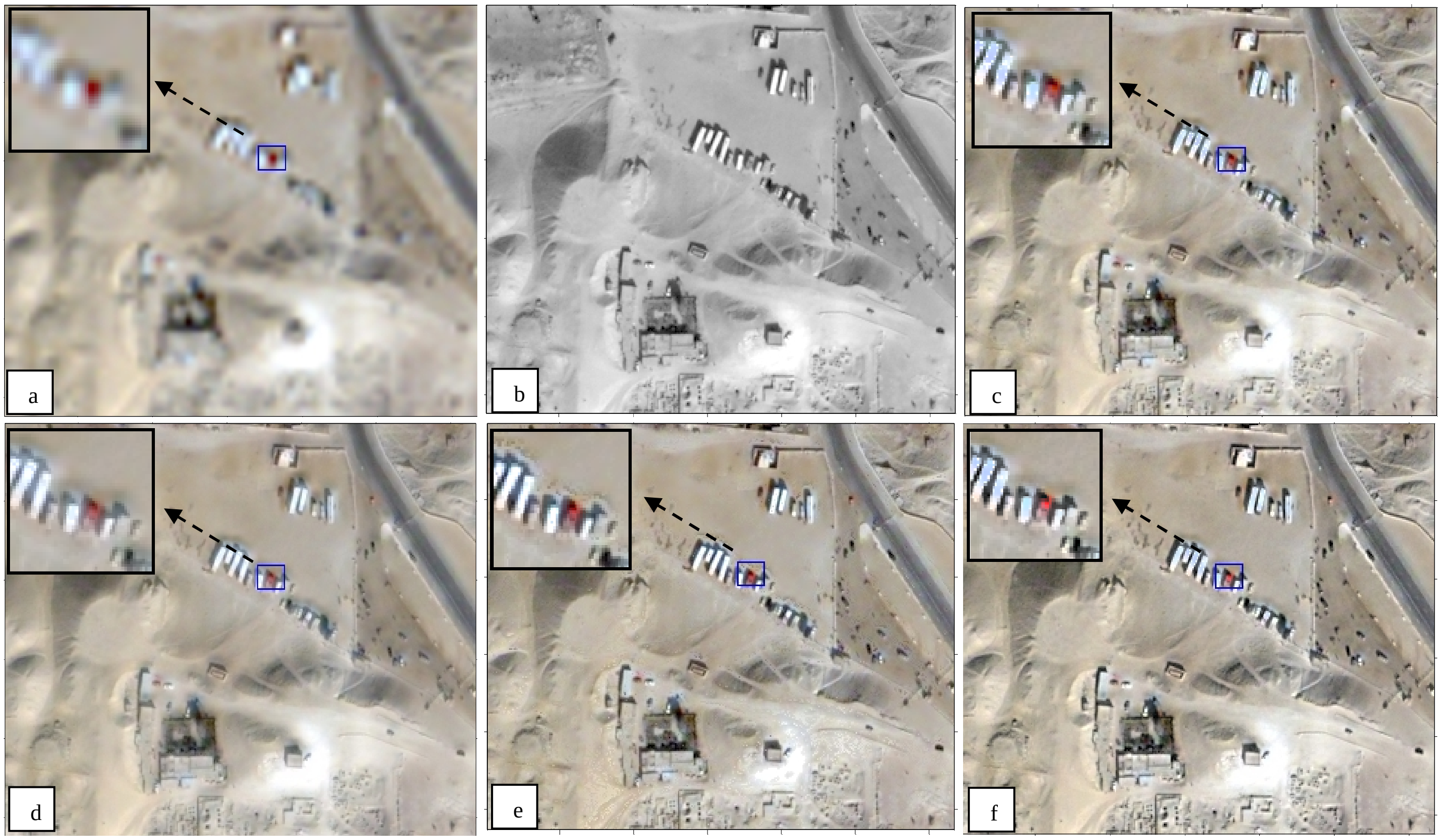

Fig. 6. (a) LMS. (b) PAN. (c) Fused image by proposed method. (d) Fused image by Wavelet Kalman Filter. (e) Fused image by Adaptive IHS. (f) Fused image by Improved Adaptive PCA.

The Table IV shows the values of CC, ERGAS, UIQI and Q4 for proposed method and some famous methods. As can be seen from this table, the quality indexes obtained by applying proposed method are pretty good compared with those obtained by applying other methods.

TABLE IV
PERFORMANCE COMPARISON WITH QUANTITATIVE INDICATORS

| | CC | ERGAS | UIQI | Q4 | SAM |
|---|---|---|---|---|---|
| IHS | 0.9261 | 3.6568 | 0.8439 | 0.8128 | 2.0514 |
| HPM | 0.8949 | 3.1529 | 0.8213 | 0.8491 | 1.9436 |
| PCA | 0.9125 | 3.9711 | 0.8444 | 0.8033 | 2.2737 |
| AW | 0.9564 | 2.8228 | 0.8937 | 0.8595 | 1.7461 |
| SW | 0.9052 | 3.4931 | 0.8519 | 0.8267 | 2.006 |
| udWI | 0.9675 | 1.5363 | 0.9292 | 0.9185 | 1.6424 |
| udWPC | 0.9741 | 1.3381 | 0.9274 | 0.9006 | 1.6799 |
| Adaptive IHS | 0.9517 | 2.4235 | 0.9265 | 0.8944 | 1.8741 |
| Improved Adaptive PCA | 0.9757 | 1.9737 | 0.9305 | 0.9016 | 1.5460 |
| Wavelet Kalman filter | 0.9701 | 1.2107 | 0.9329 | 0.9368 | 1.2441 |
| Proposed Method | **0.9862** | **1.2078** | **0.9651** | **0.9467** | **0.9080** |

Moreover, for a better comparison of the methods, for each quality metric, we have ordered the methods from the best in the upper to the worst in the lower row in Table V. The proposed method is the best in all metrics.

TABLE V
COMPARISON OF THE IMAGE FUSION METHODS

| CC | ERGAS | UIQI | Q4 | SAM |
|---|---|---|---|---|
| Proposed Method | Proposed Method | Proposed Method | Proposed Method | Proposed Method |
| Improved Adaptive PCA | Wavelet Kalman filter | Wavelet Kalman filter | Wavelet Kalman filter | Wavelet Kalman filter |
| udWPC | udWPC | Improved Adaptive PCA | udWI | Improved Adaptive PCA |
| Wavelet Kalman filter | udWI | udWI | Improved Adaptive PCA | udWI |
| udWI | Improved Adaptive PCA | udWPC | udWPC | udWPC |
| AW | Adaptive IHS | Adaptive IHS | Adaptive IHS | AW |
| Adaptive IHS | AW | AW | AW | Adaptive IHS |
| IHS | HPM | SW | HPM | HPM |
| PCA | SW | PCA | SW | SW |
| SW | IHS | IHS | IHS | IHS |
| HPM | PCA | HPM | PCA | PCA |

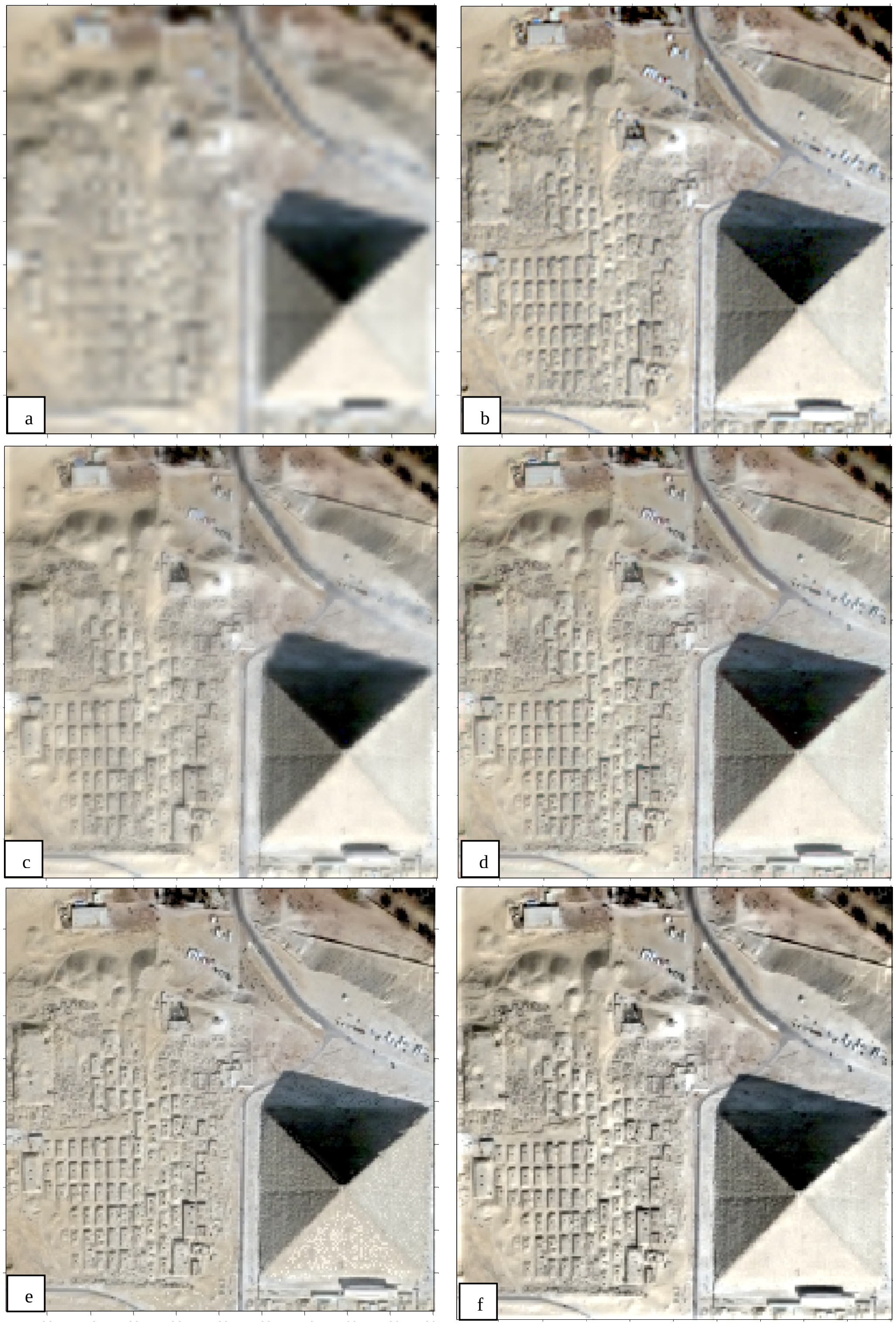

Fig. 7. (a) LMS. (b) Reference MS image. (c) Fused image by Wavelet Kalman Filter. (d) Fused image by proposed method. (e) Fused image by Adaptive IHS. (f) Fused image by Improved Adaptive PCA.

These statistical assessment findings agree with those of the visual analysis. Visual and statistical assessments show that the proposed methods have the least color distortions and desirable spatial information.

## VI. Conclusion

We studied the assumptions of recently proposed wavelet kalman filter algorithm in this paper. By taking advantage of mutual information, we showed that first order markov model is an acceptable assumption in natural MS and PAN images when ATWT is used for image decomposition. After that, we explained why posterior density in state-space model has not Gaussian distribution. Kalman filter assumes that this density has Gaussian distribution so it is not so much suitable for fusion. To overcome this deficiency, we proposed wavelet kalman particle filter method which is a more realistic model.

Then we showed that the higher frequency components of LMS have correlation with higher frequency components of reference MS images, so we exploited these higher frequency components of LMS as another observation vector, in addition to PAN, to improve our method.

Finally, the visual results show that proposed method can achieve better performance in comparison with some new methods. In addition to the visual inspection, the performance of proposed method and some former well-known methods were analyzed quantitatively. We applied correlation coefficient, ERGAS, UIQI and Q4 metrics to obtain the quality values. The performance evaluation metrics confirmed the superiority of our proposed method.